\documentclass{ecai}

\usepackage{latexsym}
\usepackage{amssymb}
\usepackage{amsmath}
\usepackage{amsthm}
\usepackage{algorithm}
\usepackage[noend]{algorithmic}
\usepackage{booktabs}
\usepackage{enumitem}
\usepackage{graphicx}
\usepackage{color}
\usepackage{xcolor}
\usepackage[caption=false,font=normalsize,labelfont=sf,textfont=sf]{subfig}
\usepackage{makecell}
\usepackage[export]{adjustbox}

\newtheorem{theorem}{Theorem}

\newtheorem{definition}{Definition}

\newcommand{\BibTeX}{B\kern-.05em{\sc i\kern-.025em b}\kern-.08em\TeX}

\begin{document}

%%%%%%%%%%%%%%%%%%%%%%%%%%%%%%%%%%%%%%%%%%%%%%%%%%%%%%%%%%%%%%%%%%%%%%%%

\begin{frontmatter}

%%% Use this command to specify the title of your paper.

\title{FedLAD: A Linear Algebra Based Data Poisoning \\ Defence for Federated Learning}

\author[A]{\fnms{Qi}~\snm{Xiong}}
\author[A]{\fnms{Hai}~\snm{Dong}\thanks{Corresponding Author. Email: hai.dong@rmit.edu.au.}}
\author[B]{\fnms{Nasrin}~\snm{Sohrabi}}
\author[A]{\fnms{Zahir}~\snm{Tari}}

\address[A]{School of Computing Technologies, Centre of Cyber Security Research and Innovation (CCSRI), RMIT University}
\address[B]{Cyber Security discipline, School of Information Technologies, Deakin University}

%%% Use this environment to include an abstract of your paper.

\begin{abstract}
Sybil attacks pose a significant threat to federated learning, as malicious nodes can collaborate and gain a majority, thereby overwhelming the system. Therefore, it is essential to develop countermeasures that ensure the security of federated learning environments. We present a novel defence method against targeted data poisoning, which is one of the types of Sybil attacks, called Linear Algebra-based Detection (FedLAD). Unlike existing approaches, such as clustering and robust training, which struggle in situations where malicious nodes dominate, FedLAD models the federated learning aggregation process as a linear problem, transforming it into a linear algebra optimisation challenge. This method identifies potential attacks by extracting the independent linear combinations from the original linear combinations, effectively filtering out redundant and malicious elements. Extensive experimental evaluations demonstrate the effectiveness of FedLAD compared to five well-established defence methods: Sherpa, CONTRA, Median, Trimmed Mean, and Krum. Using tasks from both image classification and natural language processing, our experiments confirm that FedLAD is robust and not dependent on specific application settings. The results indicate that FedLAD effectively protects federated learning systems across a broad spectrum of malicious node ratios. Compared to baseline defence methods, FedLAD maintains a low attack success rate for malicious nodes when their ratio ranges from 0.2 to 0.8. Additionally, it preserves high model accuracy when the malicious node ratio is between 0.2 and 0.5. These findings underscore FedLAD's potential to enhance both the reliability and performance of federated learning systems in the face of data poisoning attacks. For further details, we have open-sourced our work at https://gitlab.com/qinqin65/fedlad.
\end{abstract}

\end{frontmatter}

%%%%%%%%%%%%%%%%%%%%%%%%%%%%%%%%%%%%%%%%%%%%%%%%%%%%%%%%%%%%%%%%%%%%%%%%

\section{Introduction}
Federated learning (FL), proposed by \cite{mcmahan2017communication}, is a machine learning (ML) network with a star topology in which computing nodes submit their locally trained ML models to a central server to obtain a global ML model using an average merging method (referred to as FedAVG). The global ML model is then sent back to the nodes to continue their ML tasks. This approach is designed to protect the data privacy of nodes since it allows ML models to be trained without data being sent to a central server. However, this architecture is vulnerable to various attacks in which malicious nodes can attack the global model. Especially in Sybil attacks, malicious nodes can collude with each other and can easily occupy the majority, rendering FL systems unreliable~\cite{levine2006survey, 9767718}. In Sybil attacks, various fundamental attacks can be performed. According to \cite{awanCONTRADefendingPoisoning2021}, these attacks can be classified into model poisoning attacks, such as backdoor attacks \cite{bagdasaryan2021blind, bagdasaryan2020backdoor}, and data poisoning attacks, such as label flipping \cite{tolpegin2020data}. These attacks can be further classified into targeted attacks and untargeted attacks \cite{tolpeginDataPoisoningAttacks2020, sandeepaSHERPAExplainableRobust2024}. In untargeted attacks, malicious nodes aim purely to lower the performance of the global ML model \cite{tolpeginDataPoisoningAttacks2020}.  In targeted attacks, malicious nodes have specific goals for attacking the FL system~\cite{tolpeginDataPoisoningAttacks2020}. Backdoor attacks \cite{bagdasaryan2021blind, bagdasaryan2020backdoor}, which insert triggers into the ML model \cite{liuFriendlyNoiseAdversarial2022} to improve the performance of malicious labels, are a common example of such targeted attacks~\cite{tolpegin2020data, liuFriendlyNoiseAdversarial2022}. As highlighted in~\cite{tolpegin2020data}, targeted label flipping attacks are particularly effective against FL systems and can be executed even by adversaries with limited capabilities. Hence, our work focuses on defending against \textbf{targeted} data poisoning attacks.

Existing defence methods that focus on data poisoning defence in FL can be classified into the categories of {\it clustering-alike defence}, such as Sherpa~\cite{sandeepaSHERPAExplainableRobust2024}, which focuses on applying clustering techniques such as HDBSCAN to identify malicious clusters base on their ML models' interpretability; and {\it robust training-based defence}, such as the median merging method (i.e., aggregation) \cite{yin_byzantine-robust_2018, xie_generalized_2018}, which replaces the average merging method with an attack-aware merging method. All the existing methods have the limitation of only tolerating a certain proportion of malicious nodes. For example, in clustering-alike defence methods, the maximum tolerance of malicious nodes is 50\%. If the ratio of malicious nodes exceeds 50\%, the major cluster is dominated by malicious nodes, making it unreliable as an indicator of whether the cluster is benign or malicious. This limitation makes federated learning vulnerable to Sybil attacks in which malicious nodes can collude with each other and occupy the majority~\cite{levine2006survey, 9767718}. Robust training-based defence methods consider the impact of malicious ML models during the training or aggregation process. However, when the majority of ML models are malicious, these methods can be overwhelmed.

To address these limitations of existing methods, this paper proposes a Linear Algebra-based Defence (FedLAD) to defend against data poisoning attacks in FL. FedLAD models the aggregation process in FL as a linear combination. By finding the independent linear combination from the original linear combination, we can filter out redundant and malicious ML models, hence defend against malicious attacks such as data poisoning attacks. Since the malicious ratio has a low impact on the independent linear combination, FedLAD has a high tolerance for malicious nodes. Furthermore, to improve computational efficiency, we introduce a parallel optimisation algorithm based on sub-matrix splitting, which allows FedLAD to leverage parallel computing. This represents a significant advancement over existing defence approaches, whose reliance on sequential algorithms, such as clustering, limits their parallelizability.

The primary contributions of this paper are as follows: 
\begin{itemize}
    \item We propose a defence method against targeted data poisoning attacks in Federated Learning (FL), named FedLAD, which is based on linear algebra techniques. This method demonstrates a high tolerance for malicious nodes. For instance, our experiments show that FedLAD can remain effective even when 70\% of the nodes are malicious in some datasets such as AG\_NEWS.

    \item To the best of our knowledge, we are the first to leverage parallelisation to accelerate the process of finding the independent linear combination from its original linear combination. Moreover, we provide a formal mathematical proof demonstrating that this linear algebra problem can be effectively solved with parallel computing.
\end{itemize}

We conducted experiments with malicious ratios ranging from 0.2 to 0.8 using three datasets: CIFAR10, CIFAR100, and AG\_NEWS. The results demonstrate that FedLAD is more robust to attacks compared to existing methods, including Sherpa~\cite{sandeepaSHERPAExplainableRobust2024}, CONTRA~\cite{awanCONTRADefendingPoisoning2021}, Median~\cite{yin_byzantine-robust_2018}, Trimmed Mean~\cite{yinByzantineRobustDistributedLearning2018}, and Krum~\cite{blanchard_machine_2017}.

The rest of the paper is organised as follows: Section~\ref{sec:backround} introduces the background knowledge of linear algebra that is relevant to our proposed method. Section~\ref{sec:solution_details} provides solution details of FedLAD. Section~\ref{sec:experiment} discusses the various experimental results. Section~\ref{sec:related_works} summarises some of the existing works. We conclude the paper in Section~\ref{sec:conclusion}.

\section{Background}
\label{sec:backround}
Our proposed defence method is based on linear algebra. Therefore, we list all related definitions and theories in this section according to the book~\cite{axler2024linear}.

\begin{definition}
    \label{def:linear_combination}
    A \textbf{linear combination} is a set of vectors in a vector space~\cite[p.~46]{axler2024linear}. The equation below shows an example of a linear combination:
\end{definition}
\begin{equation}
    a_1v_1+a_2v_2+...+a_nv_n,
\end{equation}
\noindent where $a_i$ is a coefficient, $v_i$ is a vector.

\begin{definition}
    \label{def:span}
    A \textbf{span} is all the linear combinations of the vectors that form a subspace in a vector space~\cite[p.~47]{axler2024linear}. The equation below shows an example of a span:
\end{definition}
\begin{equation}
    \text{span}(v_1,...,v_n)=a_1v_1+a_2v_2+...+a_nv_n,
\end{equation}
\noindent where $a_1v_1+a_2v_2+...+a_nv_n$ is the set of linear combinations that spans the vectors $v_1,...,v_n$.

\begin{definition}
    \label{def:independent_linear_combination}
    A linear combination is an \textbf{independent linear combination} if the only way for the combination to equal zero is by setting all coefficients to zero~\cite[p.~50]{axler2024linear}. It means that no vector in the set can be represented by a linear combination of other vectors in the same set. The following equation is an example of an independent linear combination:
\end{definition}
\begin{equation}
    a_1v_1+...+a_nv_n=0 \iff a_1=...=a_n=0,
\end{equation}
\noindent where $v_i$ is a vector from the linear combination in the equation.

\begin{definition}
    \label{def:basis}
    A \textbf{basis} of a vector space is a set of linear independent vectors that span that vector space~\cite[p.~57]{axler2024linear}.
\end{definition}

\begin{definition}
    \label{def:rank}
    A \textbf{rank} of a matrix is the number of linearly independent vectors in the rows or columns of that matrix~\cite[p.~77]{axler2024linear}.
\end{definition}

\begin{definition}
    \label{def:independent_vectors}
    The vectors from an independent linear combination are defined as \textbf{independent vectors}. The vectors that are represented by the independent linear combination are defined as \textbf{dependent vectors}.
\end{definition}

\begin{theorem}
    \label{theorem:vectors_basis}
    Every vector space that has a finite dimension has a basis~\cite[p.~59]{axler2024linear}.
\end{theorem}

\begin{theorem}
    \label{theorem:unique_number_independent}
    The number of vectors in independent linear combinations of a vector space is unique (i.e., all bases in a vector space have the same dimension)~\cite[p.~62]{axler2024linear}. Accordingly, the rank of a matrix is unique since a matrix can be viewed as a list of vectors.
\end{theorem}

\begin{theorem}
    \label{theorem:row_col_same_rank}
    The row rank and column rank of a matrix are the same~\cite[p.~78]{axler2024linear}.
\end{theorem}

\section{The Proposed Method}
\label{sec:solution_details}

\subsection{Threat Model}
Malicious nodes engage in targeted label flipping on their local datasets, which are used to train their local machine learning models. Because these malicious models are influenced by corrupted datasets, we propose a method to detect the compromised ML models on the server side as a defence against data poisoning attacks. In light of this approach and our emphasis on mitigating targeted data poisoning attacks, we model the attacks from these malicious nodes with the following equation:
\begin{equation}
    \arg\min_{w_{m,t}}F(w_{m,t},x_{m,t},y_{m,t}),
\end{equation}
which is to train the local ML model with polluted data $x_{m,t}$ and label $y_{m,t}$ by optimising the loss function $F$. The optimised local poisoned ML models $w_{m,t}$ will be merged into the global ML model if not detected.

\subsection{Problem Definition}
In FL, every node trains its local ML model, and those local ML models are merged into a global model. The process can be formulated as:
\begin{equation}
    w_{g,t}=FedAvg(S_t)=\frac{1}{|S_t|}\sum_i w_{i,t},
\end{equation}
where $w_{g,t}$ is the global ML model parameter at the round $t$, $w_{i,t}$ is the $i$-th local ML model at the round $t$. In our case, there are malicious nodes in the system. We define the set $M$ to indicate a malicious node set. We define the set $B$ to indicate a benign node set. We define $S_t$ to indicate the whole set at the round $t$ that contains $M_t$ and $B_t$ (i.e., $S_t=M_t\cup B_t$). The problem of our work is to optimise the equation below:
\begin{equation}
    w_{b,t}-FedAvg(B_t)\rightarrow 0,
\end{equation}
where $w_{b,t}$ is the model parameter merged by nodes classified as benign. The above equation means that $w_{b,t}$ should be close to the global ML model with malicious ML models excluded so that their difference can approach 0. We get $w_{b,t}$ by the below equation:
\begin{equation}
    w_{b,t}=\frac{1}{\sum_i^{|S_t|}f^d(w_{i,t})}\sum_i^{|S_t|}w_{i,t}\times f^d(w_{i,t}),
\end{equation}
where $f^d$ is the malicious node detection function that classifies a node as either benign or malicious. The output is 0 if the node belongs to the malicious set; otherwise, the output is 1. This is represented by the following equation:
\begin{equation}
    f^d(w_{i,t})=
        \begin{cases}
            1,w_{i,t}\in B_t\\
            0,w_{i,t}\in M_t
        \end{cases}
\end{equation}

\subsection{Solution Details}
\subsubsection{Preliminary}
We model the aggregation process (i.e., merge local ML models into a global ML model) in FL as a linear combination in which the global ML model is represented by a linear combination of local ML models. The Global and Local ML models form a vector space. The equation below illustrates this concept:
\begin{equation}
    a_1w_1+...+a_nw_n=w_g \: s.t. \: \{\sum_{i=1}^n a_i=1,a_i\geq 0\},
\end{equation}
where $a_i$ is the weight of its respective local ML model. The local ML models may have different weights based on the aggregation scheme, but the sum of all the weights must equal one. $w_i$ indicates a local ML model. $w_g$ is the global ML model. From this perspective, all existing aggregation schemes such as FedAVG~\cite{mcmahan2017communication} are linear combinations. However, they do not reduce the linear combination to a simplified form so that the performance could be further optimised. We propose employing linear combination techniques to optimise the aggregation process to filter out malicious and redundant ML models.

According to Definition~\ref{def:basis} and Theorem~\ref{theorem:vectors_basis}, there exists a subset of local ML models that is a basis of the vector space of the ML models, and they are linear independent. The linear combination of the vectors from the basis can represent other vectors in this vector space. In this solution, we find out independent linear combinations of local ML models and filter out those that are not when performing aggregation in FL. The intuition is that if we can carry out a global ML model by combining only necessary local ML models, then there is no need to include other local ML models that are redundant and may be malicious.

\begin{figure}[htbp]
\centering
\includegraphics[width=0.3\textwidth]{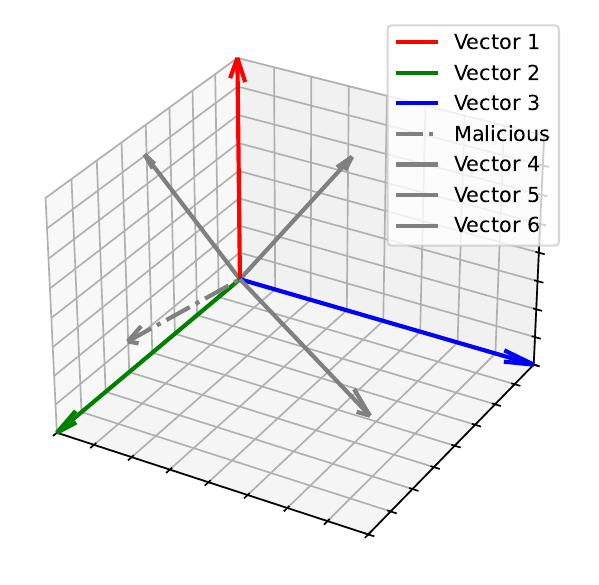}
\caption{Vectors in a 3-Dimensional Vector Space.}
\label{fig:vectors}
\end{figure}

Since the set of linear independent local ML models (i.e., vectors) is a basis, the coefficients (i.e., aggregation weights of local ML models) are the coordinates of the dependent vectors. Figure~\ref{fig:vectors} shows an example of this idea. The three vectors with colours of red, green and blue (i.e., Vector 1, 2, and 3) form a basis of the three-dimensional vector space. The grey vectors (i.e., Vector 4, 5, 6 and Malicious)  are dependent vectors that can be represented by the linear combination of the three base vectors (i.e., Vector 1, 2, and 3). The coordinates (i.e., the coefficients of the linear combination) in the vector space can uniquely identify a grey vector. One of the grey vectors is malicious, but we only consider independent vectors, so the malicious vector will be filtered out during the aggregation process.

If a dependent vector can be represented by a linear combination, then it means the information of the dependent vector is contained in the basis and coordinate system. Hence, the dependent vectors can be regarded as redundant vectors that may contain malicious data and should be filtered out. We provide a mathematical proof for the idea that the dependent linear combinations are redundant and can be safely discarded:
\begin{proof}
In federated learning, the global model $w_g$ is typically computed as a weighted average (linear combination) of local models $w_1,w_2,...,w_n$:
\begin{equation}
    w_g=\sum_{i=1}^n a_iw_i, \text{where} \sum_{i=1}^n a_i=1,a_i\geq 0,
\end{equation}
which forms a vector space $V\subseteq \mathbb{R}^d$, where $w_i\in \mathbb{R}^d$ is a flattened model parameter vector. A set of vectors $\{w_1,...,w_n\}$ is linearly dependent if there exists a nontrivial combination such that:
\begin{equation}
    \sum_{i=1}^n \alpha_iw_i=0 \; \text{with some} \; \alpha_i\neq 0,
\end{equation}
which implies that at least one vector $w_j$ can be written as a linear combination of the others:
\begin{equation}
    w_j=\sum_{i\neq j} \beta_iw_i.
\end{equation}

Hence, $w_j$ adds no new direction or information to the span of the set. Including it in the aggregation does not expand the representational capacity of the model space. According to the definition~\ref{def:basis}, a basis of a vector space is a minimal set of linearly independent vectors that span the space. According to the theorem~\ref{theorem:vectors_basis} and~\ref{theorem:unique_number_independent}, any vector in the space can be uniquely represented as a linear combination of the basis vectors. Therefore, if we identify a basis $\{v_1,...,v_k\}\subseteq \{w_1,...,w_n\}$, then:
\begin{equation}
    \text{span}(w_1,...,w_n)=\text{span}(v_1,...,v_k),
\end{equation}
which means the global model $w_g$ can be constructed entirely from the basis vectors. The remaining dependent vectors are redundant. By discarding linearly dependent models, the span of the model space remains unchanged, and no information is lost. Noise and malicious influence are reduced (i.e., dependent vectors may be adversarial replicas or perturbations of benign models). Aggregation over fewer, independent vectors reduces complexity, hence the computation is optimised. This is the core idea behind FedLAD, which utilises the row-reduced echelon form (RREF) to extract the pivot rows (basis vectors) and discard the rest.

In conclusion, mathematically, linearly dependent local models do not contribute new information to the global model. Therefore, they can be safely discarded without affecting the representational power of the aggregation. This not only improves robustness against poisoning attacks but also enhances computational efficiency.
\end{proof}

This idea is also backed up by the row reduced echelon form (RREF) of a matrix since a matrix can be viewed as a set of vectors. The RREF of a matrix is a form that only keeps the pivot rows of the matrix while other rows are zeroed out~\cite[p.~30]{manglik2024linear}. The pivot rows contain all the necessary information in a matrix to solve a linear algebra problem~\cite[p.~30]{manglik2024linear}.

\subsubsection{Solution Overview}
\begin{figure*}[htbp]
\centering
\includegraphics[width=0.8\textwidth]{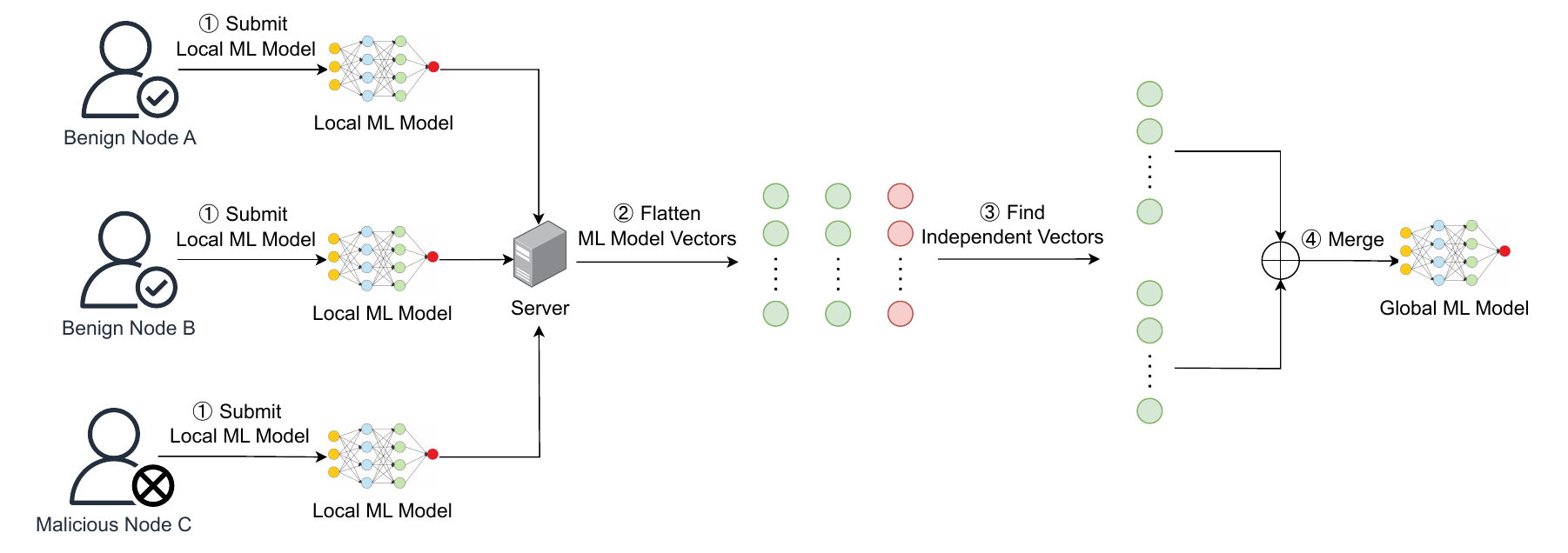}
\caption{The overview of FedLAD.}
\label{fig:overview}
\end{figure*}

Based on the above foundation, in FedLAD, we organise local ML models into a matrix and compute its RREF. The list of pivot rows (also simplified as pivot) of the RREF is the independent linear combination of the local ML models. We only select local ML models corresponding to these pivot rows for the aggregation process in FL. Hence, the impact of malicious ML models caused by data poisoning attacks can be eliminated. 

Figure~\ref{fig:overview} shows the overview of the process of FedLAD. There are three nodes: A, B, and C, with node C being malicious. The circled numbers indicate the step of the process. In the first step, the nodes submit their local ML models to a server. In the second step, the server flattens the ML models into vectors and lists them as a matrix. In the third step, the server finds the independent ML model vectors by calculating the RREF of the matrix and then selects only local ML models from the pivot rows in the RREF. In the last step, the selected local ML models are merged into a global ML model. The following sections cover technical details in the process. 

\subsubsection{Flatten ML models}
ML models may have multiple layers, and we need to flatten them to one-dimensional (1-D) vectors. In FedLAD, we flatten ML models into 1-D vectors based on row-major (C-style) order~\cite[p.~253]{manglik2024introduction}.

\subsubsection{Calculate RREF}
We apply Gaussian Elimination (GE)~\cite[p.~30]{manglik2024linear} to calculate the RREF of a matrix. Algorithm~\ref{alg:calc_rref} shows the detailed pseudo code of calculating RREF for a given matrix. It finds possible pivots by searching from each column and row. The ``cross cancel'' function in Algorithm~\ref{alg:cross_cancel} zeros out other elements except the pivot in a row to make sure only the pivot is non-zero based on the definition of RREF~\cite[p.~30]{manglik2024linear}. According to~\cite[p.~233]{manglik2024linear}, the RREF for a matrix is unique, so there is no need to handle possible multiple forms of independent linear combinations.

This paragraph details how RREF is calculated in Algorithm~\ref{alg:calc_rref}. Lines~\ref{alg_line:for_col} and~\ref{alg_line:for_row} iterate through the columns and rows. Line~\ref{alg_line:row_col} retrieves the rows of the current column that are below the currently processed row. Line~\ref{alg_line:possible_pivot_index} finds the index of the maximal number regardless of the number sign (i.e., positive or negative) from the rows obtained from line~\ref{alg_line:row_col}. Line~\ref{alg_line:abs_possible_pivot} determines whether the index in line~\ref{alg_line:possible_pivot_index} is a possible pivot index. The number in the index must not be zero otherwise, its rank will be zero. Lines from~\ref{alg_line:t_row} to~\ref{alg_line:row_t} swap the row from the index in line~\ref{alg_line:pivot_index} with the current row. Lines from~\ref{alg_line:row_above} to~\ref{alg_line:cross_cancel_row_below} zero out the rows except the current row to ensure only the pivot row is non-zero, which is a characteristic of RREF~\cite[p.~30]{manglik2024linear}. The steps from~\ref{alg_line:t_row} to~\ref{alg_line:row_t} and from~\ref{alg_line:row_above} to~\ref{alg_line:cross_cancel_row_below} are the GE steps to eliminate unnecessary rows. In line~\ref{alg_line:row_index+1}, we continue to search for other possible pivots after the current row. In line~\ref{alg_line:early_stop}, if no possible pivot is found, we perform an early stop to conserve computational resources and move on to the next column. After finishing the iteration of columns and rows, the found pivots are returned.

\begin{algorithm}[!htb]
\caption{Calculate RREF}
\label{alg:calc_rref}
\textbf{Input}: Local ML Models $W=[w_1,w_2,...,w_n]^T$\\
\textbf{Output}: RREF pivots
\begin{algorithmic}[1]
\STATE let $\text{pivot\_cols}=[\:]$
\STATE let $\text{row\_index}=0$
\STATE let $\text{row\_size}$ be the number of rows of $W$
\STATE let $\text{col\_size}$ be the number of columns of $W$
\FOR {$col$ in $\{0,..., \text{col\_size}\}$} \label{alg_line:for_col}
\FOR {$row$ in $\{\text{row\_index},...,\text{row\_size}\}$} \label{alg_line:for_row}
% \STATE \textcolor{blue}{// get the rows of the current column in the iteration}
\STATE let $\text{row\_col}=W[row:,col]$ \label{alg_line:row_col}
% \STATE \textcolor{blue}{// find the possible pivot index whose value must not be zero}
\STATE let $\text{possible\_pivot}=argmax(abs(\text{row\_col}))$ \label{alg_line:possible_pivot_index}
\IF {$abs(\text{row\_col}[\text{possible\_pivot}]) > 0$} \label{alg_line:abs_possible_pivot}
\STATE let $\text{pivot}=\text{possible\_pivot}$ \label{alg_line:pivot_index}
% \STATE \textcolor{blue}{// add the pivot to the list}
\STATE add $col$ to $\text{pivot\_cols}$
% \STATE \textcolor{blue}{// exchange the pivot row with the current row based on GE operations}
\IF {$\text{pivot}\neq 0$}
\STATE let $t = W[row]$ \label{alg_line:t_row}
\STATE let $W[row]=W[row+\text{pivot}]$
\STATE let $W[row+\text{pivot}]=t$ \label{alg_line:row_t}
\ENDIF
\STATE let $\text{row\_above}=W[:row]$ \label{alg_line:row_above}
\STATE let $\text{row\_below}=W[row+1:]$
\STATE let $\text{row\_col}=W[row,col]$
% \STATE \textcolor{blue}{// zero out rows above the pivot}
\IF {the row size of $\text{row\_above}>0$}
\STATE $W[:row]=\text{cross\_cancel(row\_col,}$
$\text{$W[row]$, row\_above, col)}$
\ENDIF
% \STATE \textcolor{blue}{// zero out rows below the pivot}
\IF {the row size of $\text{row\_below}>0$}
\STATE $W[row+1:]=\text{cross\_cancel(row\_col,}$ \label{alg_line:cross_cancel_row_below}
$\text{$W[row]$, row\_below, col)}$
\ENDIF
\STATE $\text{row\_index }+=1$ \label{alg_line:row_index+1}
\ELSE
% \STATE \textcolor{blue}{// early stop if no pivot found}
\STATE break \label{alg_line:early_stop}
\ENDIF
\ENDFOR
\ENDFOR
\RETURN \text{pivot\_cols}
\end{algorithmic}
\end{algorithm}

\subsubsection{Global ML Model Aggregation}
After the RREF of the matrix of local ML models is carried out, we apply FedAVG on the local ML models that are the RREF pivots (i.e., same aggregation weight). The rationale is that information from the independent vectors is considered equally important. By doing so, we can limit the influence of malicious ML models while ensuring that valuable information from normal ML models is not overlooked.

\begin{algorithm}[htb]
\caption{Cross Cancel}
\label{alg:cross_cancel}
\textbf{Input}: Current Value $v$, Current Row $row$, Considered Rows $\text{rows}$, Current Column $col$\\
\textbf{Output}: Cross Cancelled Rows
\begin{algorithmic}[1]
\STATE repeat $\text{rows}[:,col]$ horizontally and turn them into a square matrix and set it as $a$
\STATE repeat $row$ horizontally and turn them into a matrix with the column size equal to the row size of $a$ and set it as $b$
\STATE let $c=a\times b$
\RETURN $v\times \text{rows}-c$
\end{algorithmic}
\end{algorithm}

\subsubsection{Computation Optimisation}
\label{sec:computation_opt}
As the size of a matrix increases, the time required for computation also increases. Algorithm~\ref{alg:calc_rref} addresses this issue by allowing the algorithm to break out early if it determines that a row is certainly not a pivot. This helps avoid unnecessary calculations. However, this algorithm serially iterates through the columns and rows without parallelism, resulting in wasted idle computing resources. To improve efficiency and reduce computation time, we propose a parallel version to perform this task.

Algorithm~\ref{alg:parallel_rref} outlines the details of this approach. It involves splitting the given matrix into sub-matrices. The example in Algorithm~\ref{alg:parallel_rref} illustrates a column-wise split, but a row-wise split would yield the same result~\cite[p.~123]{gentle2024matrix}. Once the matrix is divided into sub-matrices, multiple computing units (CPUs) can process different sub-matrices simultaneously. After calculating the RREF for each sub-matrix, we combine these results into a new matrix and repeat the process of splitting and combining until we are left with the smallest-sized sub-matrix. The RREF from this smallest sub-matrix is equivalent to that of the original matrix. We provide a proof below. 

\begin{proof}
According to~\cite[p.~48]{manglik2024linear}, the sum of the rank of sub-matrices is greater than or equal to that of the original matrix, which is:
\begin{equation}
    Rank(M_1)+Rank(M_2)+...+Rank(M_n)\geq Rank(M_o),
\end{equation}
where $M_i$ is a sub-matrix of the original matrix $M_o$, $Rank$ is a function that indicates the rank of a matrix. We calculate the RREF of each sub-matrix. Then take the non-zero vectors from the matrices and combine them into a new matrix $M_N$. The rest zero vectors are put to a matrix $M_z$. The rank of the matrix $M_N$ is equal to that of the original matrix, which is:
\begin{equation}
    Rank(M_N)=Rank(M_o),
\end{equation}
since the rank of one sub-matrix is smaller than or equal to that of the original matrix~\cite[p.~44]{manglik2024linear}, which is
\begin{equation}
    Rank(M_N) \leq Rank(M_o),
\end{equation}
but the sum of the rank of all sub-matrices is greater than or equal to that of the original matrix, which is
\begin{equation}
    Rank(M_N) + Rank(M_z) \geq Rank(M_o),
\end{equation}
and because
\begin{equation}
    Rank(M_z)=0,
\end{equation}
hence
\begin{equation}
    Rank(M_N)=Rank(M_o).
\end{equation}
We continue to split $M_N$ into sub-matrices and perform the same operations aforementioned until we get the smallest-sized matrix $M_s$. The rank of $M_s$ is the same as that of $M_o$, which is the same case as $M_N$. According to~\cite[p.~233]{manglik2024linear}, the RREF for a matrix is unique, so the RREF of $M_s$ is the same as that of the original matrix $M_o$.
\end{proof}

The sub-matrix is created by recombining the non-zero columns or rows of its parent sub-matrix after performing the RREF calculation. The root sub-matrix is obtained in the same way from the original matrix. The smallest-sized sub-matrix is defined as one that cannot be further divided. Its size is determined based on the number of available CPUs. For FedLAD, the result of the parallel RREF calculation is the same as that of the non-parallel version since the same RREF form results in the same pivots according to the definition of RREF~\cite[p.~233]{manglik2024linear}.

\begin{algorithm}[htbp]
\caption{Parallel RREF Calculation}
\label{alg:parallel_rref}
\textbf{Input}: Local ML Models $W=[w_1,w_2,...,w_n]^T$\\
\textbf{Output}: RREF pivots
\begin{algorithmic}[1]
\STATE let $n$ be the number of CPUs
\STATE let $\text{col\_size}$ be the number of columns of $W$
\STATE let $\text{smallest\_submatrix\_size}=\text{col\_size}\div n$
\STATE split $W$ into $n$ sub matrices based on the split size smallest\_submatrix\_size and assign them to sub\_matrices$[\:]$
\STATE spawn $n$ CPU threads to call Algorithm~\ref{alg:calc_rref} and pass on the respective input of sub\_matrices$[i]$ with $i\in n$
\STATE combine the returned RREFs to a new matrix $W_s$
\IF{the size of $W_s>$ smallest\_submatrix\_size}
\STATE call Algorithm~\ref{alg:parallel_rref} and pass on the input of $W_s$
\ELSE
\STATE call Algorithm~\ref{alg:calc_rref} and pass on the input of $W_s$
\ENDIF
\RETURN the RREF returned form calling Algorithm~\ref{alg:calc_rref}
\end{algorithmic}
\end{algorithm}

\section{Complexity Analysis}
\label{sec_fedlad:complexity_analysis}
We analyse the time complexity for both the serial and parallel versions of RREF calculation. For the serial version, the time complexity is 
\begin{equation}
    O(nm), 
\end{equation}
where $n$ is the number of nodes and $m$ is the number of columns in the matrix (Since our algorithm splits the matrix by columns, the complexity would be equivalent if we split by rows). In the parallel version, we model the operation as a balanced c-ary tree (where c is the number of CPUs), with splitting and combining at each level. Its complexity is 
\begin{equation}
    O((log_c^m) * n + (log_c^m) * c)=O((log_c^m) * (n+c)),
\end{equation}
in which the first term on the left side of the equation captures the computation time, while the second represents communication overhead. This demonstrates that while the serial version scales poorly, the parallel version is much more efficient and suitable for large-scale FL tasks. Since the time complexity introduced by the communication overhead is 
\begin{equation}
    O((log_c^m) * c), 
\end{equation}
which has a marginal effect with the increase of the number of CPUs, or the column size of the matrix. Therefore, the cost of splitting and combining sub-matrices does not outweigh the benefits of parallelisation, even at scale.

\section{Experiment Evaluation}
\label{sec:experiment}

\subsection{Experiment Settings}
We conduct experiments with 10 nodes, which are initialised with an array of local ML models, datasets, and node ID, to evaluate our proposed method FedLAD. The ratio of malicious nodes ranges from 20\% to 80\% (we also use floating numbers 0.2 to 0.8 to indicate the ratio), which is to verify the robustness of the proposed defence method and baseline defence methods. The malicious attack performs label flipping on the experimental datasets. The details of the label flipping are introduced hereinafter. We demonstrate the improvements of FedLAD by comparing it with five baseline methods: Sherpa~\cite{sandeepaSHERPAExplainableRobust2024}, CONTRA~\cite{awanCONTRADefendingPoisoning2021}, Median~\cite{yin_byzantine-robust_2018}, Trimmed Mean~\cite{yinByzantineRobustDistributedLearning2018}, and Krum~\cite{blanchard_machine_2017}. The justification for choosing these baseline methods is provided in Section~\ref{sec:baselines}. We conduct the experiments five times and adopt the average of the results. We also disclose the running time of FedLAD and baselines in Section~\ref{sec:running_time}. We run FedLAD on serial and parallel versions and list the respective running times to demonstrate the advantage of parallel optimisation proposed in Section~\ref{sec:computation_opt}.

\subsubsection{Computational and Software Setups}
The experiments are carried out on a high-performance computer with 64GB of memory, 16 CPUs and an A40 GPU to accelerate multiple-node federated learning with large-scale neural networks. The implementations of FedLAD and baseline methods are based on Pytorch 2.3.1, Numpy 1.26.3, Flower 1.15.0 (a federated learning framework) and Python 3.10.12.

\subsubsection{Implementation Details}
To evaluate FedLAD in various scenarios, we use two image datasets, CIFAR10 and CIFAR100,~\footnote{\label{footnote_cifar}https://www.cs.toronto.edu/\~kriz/cifar.html} and an NLP dataset, AG\_NEWS.~\footnote{http://groups.di.unipi.it/\~gulli/AG\_corpus\_of\_news\_articles.html} To enable non-IID distribution in the datasets, we apply a Dirichlet distribution~\cite{li2021model, han2022fedx} with the heterogeneity parameter set to 0.5 to sample different data distributions for each node. Following the work~\cite{awanCONTRADefendingPoisoning2021}, we use ResNet18 for the image recognition tasks on the CIFAR10 and CIFAR100 datasets. For the AG\_NEWS dataset, we use a three-layer neural network, which is the optimal structure according to our experiment, to perform the text classification task. The input layer of this neural network has 95,811 units, corresponding to the number of tokens in the AG\_NEWS dataset. The hidden layer contains 64 units, which embeds the input text into a vector space, and the output layer has four units, corresponding to the four classes. The work~\cite{awanCONTRADefendingPoisoning2021} used 300 rounds for global communication and one round for local training. We adopt the same strategy but with some adaptations based on our experimental observation, in which the ML model performance is optimal. For the CIFAR10 and CIFAR100 datasets, we trained for two rounds in the local environment and conducted 40 rounds of global communication. For the AG\_NEWS dataset, we trained for two local rounds and conducted 20 rounds of global communication. During the training, malicious nodes perform label-flipping attacks on their local dataset.

\begin{figure*}[!htbp]
    \centering
    \begin{adjustbox}{minipage=\linewidth,scale=0.8}
    \captionsetup[subfloat]{labelfont=scriptsize,textfont=scriptsize}
    \subfloat[]{\includegraphics[width=0.3\linewidth]{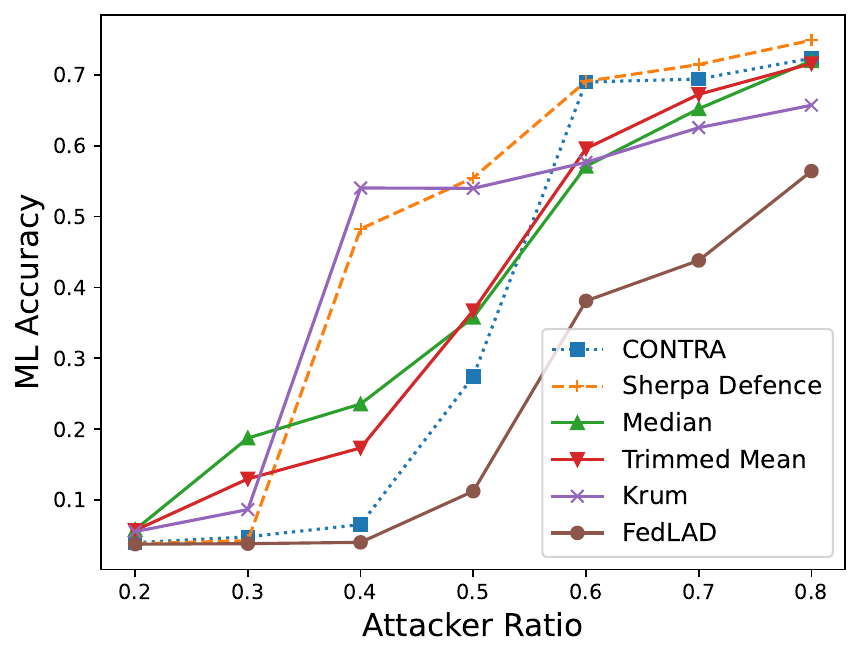} \label{fig:malicious_ratio_vs_attack_success_rate_cifar10}}
    \hfill
    \subfloat[]{\includegraphics[width=0.3\linewidth]{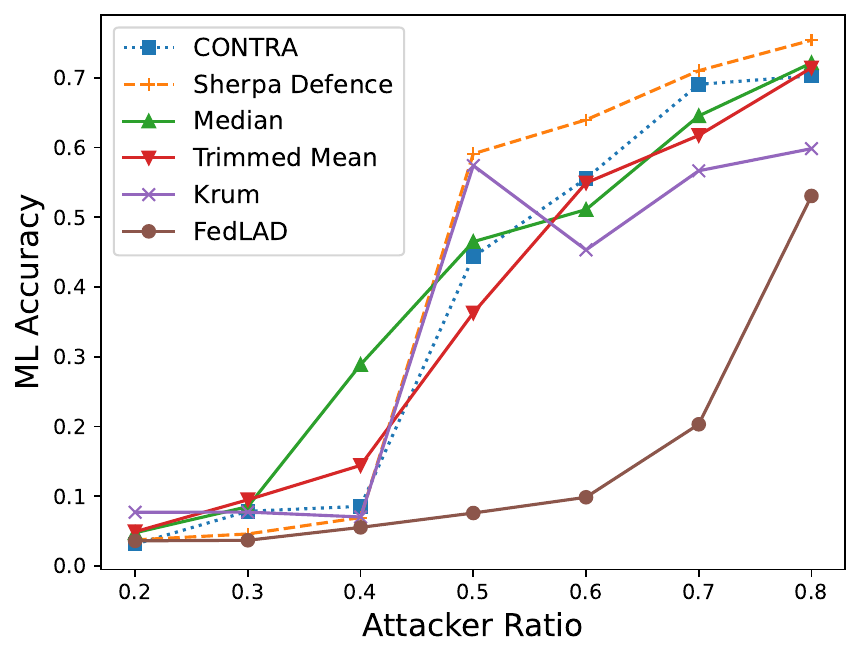} \label{fig:malicious_ratio_vs_attack_success_rate_cifar100}}
    \hfill
    \subfloat[]{\includegraphics[width=0.3\linewidth]{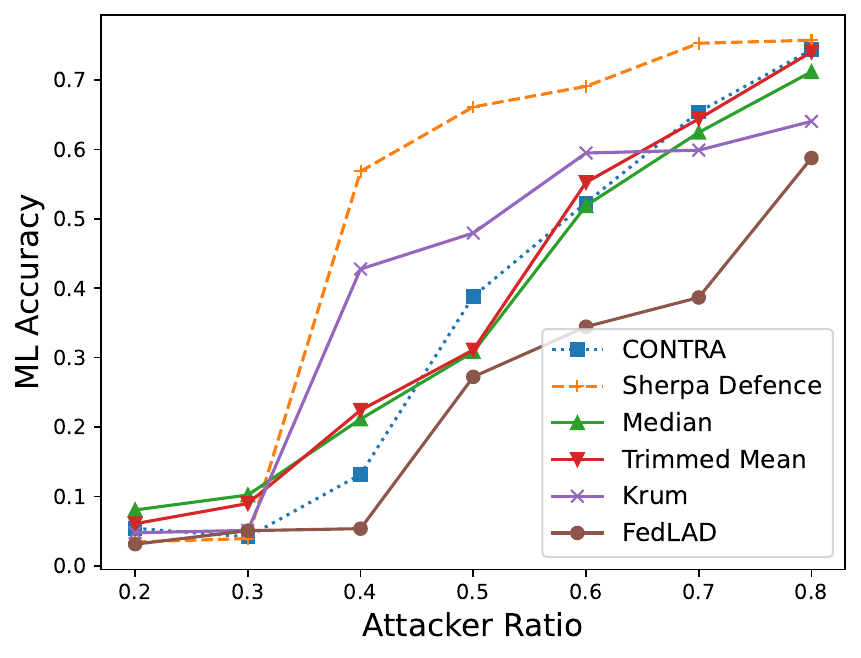} \label{fig:malicious_ratio_vs_attack_success_rate_ag_news}}
    \caption{Malicious ratio VS Attacking Success Rate on Non-IID datasets. \ref{fig:malicious_ratio_vs_attack_success_rate_cifar10} is under CIFAR10. \ref{fig:malicious_ratio_vs_attack_success_rate_cifar100} is under CIFAR100. \ref{fig:malicious_ratio_vs_attack_success_rate_ag_news} is under AG\_NEWS}.
    \label{fig:malicious_ratio_vs_attack_success_rate}
    \end{adjustbox}
\end{figure*}
\begin{figure*}[!htbp]
    \centering
    \begin{adjustbox}{minipage=\linewidth,scale=0.8}
    \captionsetup[subfloat]{labelfont=scriptsize,textfont=scriptsize}
    \subfloat[]{\includegraphics[width=0.3\linewidth]{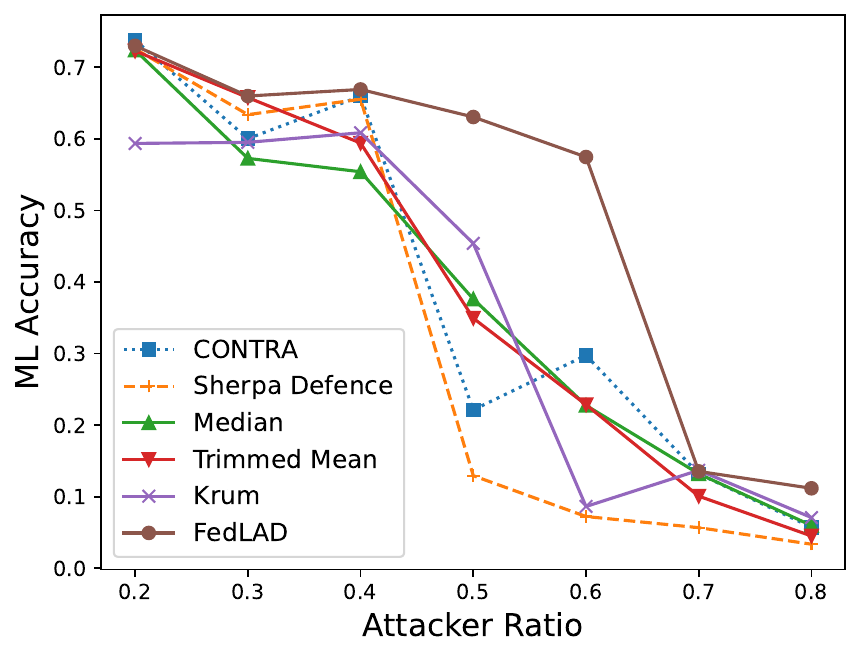} \label{fig:malicious_ratio_vs_model_accuracy_cifar10}}
    \hfill
    \subfloat[]{\includegraphics[width=0.3\linewidth]{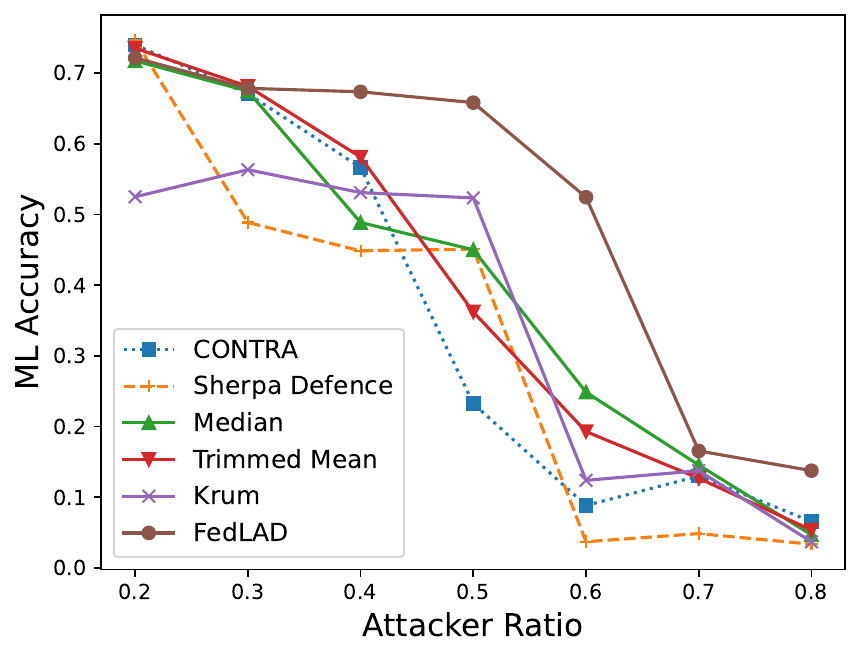} \label{fig:malicious_ratio_vs_model_accuracy_cifar100}}
    \hfill
    \subfloat[]{\includegraphics[width=0.3\linewidth]{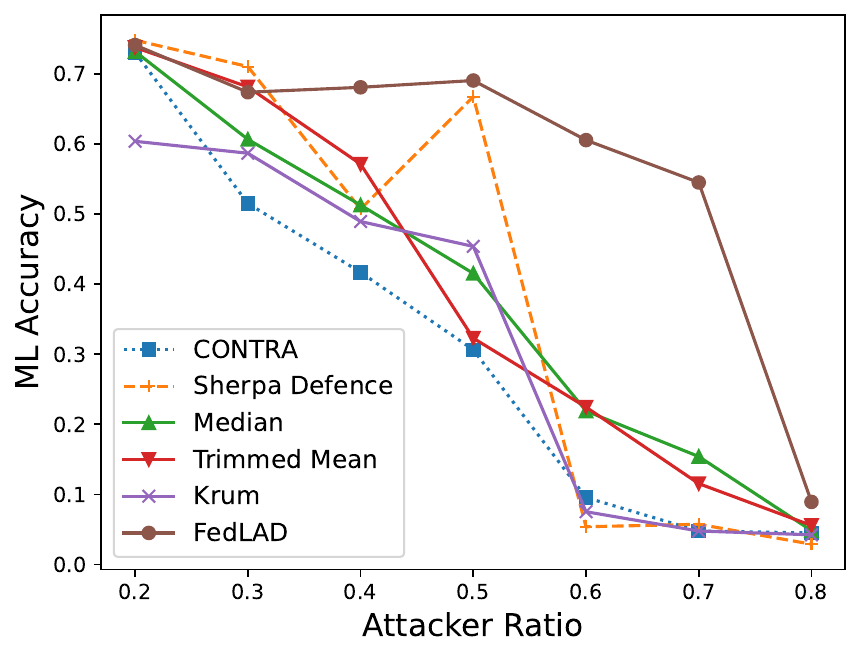} \label{fig:malicious_ratio_vs_model_accuracy_ag_news}}
    \caption{Malicious ratio VS Model Accuracy on Non-IID datasets. \ref{fig:malicious_ratio_vs_model_accuracy_cifar10} is under CIFAR10. \ref{fig:malicious_ratio_vs_model_accuracy_cifar100} is under CIFAR100. \ref{fig:malicious_ratio_vs_model_accuracy_ag_news} is under AG\_NEWS}.
    \label{fig:malicious_ratio_vs_model_accuracy}
    \end{adjustbox}
\end{figure*}

\subsubsection{Malicious Nodes Generation}
\label{sec:malicious_nodes_generation}
In the experiment setting, the malicious ratios range from 0.2 to 0.8. For each malicious ratio, we randomly draw the number of malicious nodes based on the calculation of the total number of nodes with the respective ratio. The process can be represented as:
\begin{equation}
    M=rand(S, |S|\times r),
\end{equation}
where $M$ is the malicious node set, $S$ is the whole set that contains the malicious node set and the benign node set, and $r$ is the malicious ratio. $rand$ is a random sampling function with the first parameter being the set to sample from, and the second parameter being the number of subsets to sample.

\subsubsection{Malicious Nodes Training}
When malicious nodes train their local machine learning (ML) models, they perform label flipping on the training data. Following~\cite{sandeepaSHERPAExplainableRobust2024}, we swap two targeted labels of the training dataset.

\subsubsection{Performance Metrics}
To evaluate FedLAD in terms of the success rate of attacks by malicious nodes, we utilise the attack success rate (ASR) to measure the effectiveness of a defence method in preventing successful attacks by malicious nodes. A higher ASR indicates lower performance in defending against attacks on the ML model. The ASR is calculated using the following equation:
\begin{equation}
    ASR=\frac{|\{y_i\|y_i=F(\frac{1}{|S|}\sum_i w_i,x_m), y_i\in y_m\}|}{|y_m|},
\end{equation}
where $F$ is the loss function, $S$ is the set of all nodes, $w$ is the ML model weight, and $y_m$ is the labels from malicious dataset $x_m$.

To evaluate FedLAD in terms of the impact of malicious nodes on model performance, we employ model accuracy (MA) as the metric to assess how effectively a defence method prevents malicious nodes from degrading model performance. A higher MA with certain ratios of malicious nodes indicates better performance of the defence method in maintaining model accuracy during machine learning tasks. The MA is calculated using the following equation:
\begin{equation}
    MA=\frac{|\{y_i\|y_i=F(\frac{1}{|S|}\sum_i w_i,x), y_i\in y\}|}{|y|},
\end{equation}
where $y$ is the label from the dataset $x$.

\subsubsection{Baseline Methods}
\label{sec:baselines}
In this section, we detail the baseline methods and provide justifications for selecting them to compare with FedLAD. Below, we provide a summary of each method along with our assessments.

\begin{itemize}
    \item CONTRA~\cite{awanCONTRADefendingPoisoning2021} is from the clustering-alike defence group, which relies on the cosine similarity between malicious and benign nodes. 

    \item Sherpa~\cite{sandeepaSHERPAExplainableRobust2024} is from the clustering-alike defence group, which relies on explainable AI techniques to investigate the difference of ML model interpretations between malicious and benign nodes and distinguish them.

    \item Median~\cite{yin_byzantine-robust_2018} is from the robust training-based defence group and is considered a classic defence method against malicious attacks in federated learning (FL). It has been used as a baseline for comparison in various studies~\cite{sandeepaSHERPAExplainableRobust2024}. Instead of using average aggregation (i.e., merging), it aggregates the local machine learning (ML) models based on the median value. This approach is effective against model poisoning attacks, as malicious ML models tend to deviate from the median value of the local models to some extent, as discussed in the work by~\cite{yin_byzantine-robust_2018}.

    \item Trimmed Mean~\cite{yinByzantineRobustDistributedLearning2018} is from the robust training-based defence group, which defines a trim parameter $k$. It selects only the smallest and largest $n-2k$ values from ML models. After the selection based on the trim parameter, a mean-based aggregation scheme is used for the selected ML model weights.

    \item Krum~\cite{blanchard_machine_2017} belongs to the robust training-based defence group, which is considered a classic approach and is often used as a baseline for comparison in various studies~\cite{sandeepaSHERPAExplainableRobust2024}. This method identifies the local machine learning (ML) models that are $n-f-2$ closest to the global ML model based on squared sum distance. The underlying premise of this work is that malicious ML models will show significant deviation from the global ML model in terms of their squared sum distance.
\end{itemize}

\subsection{Experiment Results and Evaluation}
\label{sec:exp_eval}
\subsubsection{Attacking Success Rate (ASR)}
Figure~\ref{fig:malicious_ratio_vs_attack_success_rate} shows the ASR for malicious ratios ranging from 0.2 to 0.8 for the CIFAR10, CIFAR100, and AG\_NEWS datasets. Compared with baseline methods, our method, FedLAD, exhibits the highest performance in resisting malicious nodes across various ratio settings, showcasing the advantage of resisting attacks from malicious nodes. The ASR of FedLAD increases sharply when the malicious ratio exceeds 0.5 on CIFAR10 and AG\_NEWS, 0.7 on CIFAR100, as there are too few benign ML models to cancel out the attacking effect of malicious ML models. The reason why FedLAD maintains a lower ASR until the malicious ratio surpasses 0.7 is that the CIFAR100 dataset has 100 labels, which limits the impact of label flipping attacks from malicious nodes.

All baseline methods show a high ASR when the malicious ratio exceeds 0.5 (i.e., ASR increases sharply). This is because these methods rely on the majority to be benign nodes, which fail when the malicious ratio surpasses 0.5. Among the baseline methods, Sherpa performs the worst probability due to that the calculation of SHAP features is based on sampling methods, which can result in some important features being omitted. Krum performs reasonably better than other baseline methods probability due to the effectiveness of excluding thirty per cent of nodes (i.e., a Byzantine tolerant setting in Krum). However, the MA performance of Krum is poorer than other baseline methods as we can see in Section~\ref{sec:ma}.

\subsubsection{Model Accuracy (MA)}
\label{sec:ma}
Figure~\ref{fig:malicious_ratio_vs_model_accuracy} shows the MA with malicious ratios ranging from 0.2 to 0.8 for the CIFAR10, CIFAR100, and AG\_NEWS datasets. Compared with baseline methods, FedLAD maintains the highest performance across all malicious ratio settings, demonstrating that FedLAD can lower the impact of malicious attacks on the MA. Performance of FedLAD drops sharply when the malicious ratio is 0.7 and 0.8. This is because there are not enough benign nodes to contribute to the MA performance. Despite this, FedLAD can still maintain high performance even when the ratio is as high as 0.8, as only the pivot ML models are selected, which can exclude redundant and malicious ML models, even at high malicious ratios.

Among the baseline models, Krum performs the worst as it maintains a low MA across the three datasets probably due to the thirty per cent of nodes are excluded. Some of these excluded nodes may contain the most important contribution to MA. FedLAD can avoid this situation since the most important contributions are always included in the RREF pivots according to the mechanism of independent linear combination in linear algebra. Sherpa slightly outperforms other baseline methods under the AG\_NEWS dataset, but with a high variance. The reason may be that the texts are encoded into vectors by an encoder. Some important information is retained, and the noise is ignored. Compared with the pixel features in pictures, the vector features are more efficient as an input to produce the SHAP~\cite{NIPS2017_7062} features.

\subsubsection{Running Time of Defence Methods}
\label{sec:running_time}

Table~\ref{tab:running_time} presents the running times of the defence methods on the CIFAR10, CIFAR100, and AG\_NEWS datasets. The running time is recorded during the five runs of the experiments, and we adopt the average value. The serial version of FedLAD has the highest running time. However, when parallel optimisation is applied, there is a significant reduction in running time, demonstrating the effectiveness of this approach. Although the running time of the parallel optimised FedLAD does not exceed that of the baseline methods, this is primarily due to our limited computational resources for conducting large-scale experiments. In fact, the parallel version is more scalable than the others (see Section~\ref{sec_fedlad:complexity_analysis}).

\begin{table}[!htbp]
    \centering
    \small
    \caption{Running Time (Minutes) of Defence Methods on Three datasets}
    \begin{tabular}{m{8 em}|c|c|c}
    \hline
   \textbf{Method} & \textbf{CIFAR10} & \textbf{CIFAR100} & \textbf{AG\_NEWS} \\
    \hline
    \hline
    \textbf{FedLAD Serial} &\textbf{ 245.63} & \textbf{281.44} & \textbf{127.51} \\
    \textbf{FedLAD} \textbf{Parallel} & \textbf{63.43} & \textbf{73.88} & \textbf{36.7} \\
    Sherpa & 58.05 & 63.45 & 35.97 \\
    CONTRA & 49.01 & 67.62 & 25.72 \\
    Median & 44.14 & 47.57 & 27.48 \\
    Trimmed Mean & 44.34 & 72.05 & 29.43 \\
    Krum & 45.51 & 68.72 & 26.62 \\
    \hline
    \end{tabular}
    \label{tab:running_time}
\end{table}

\section{Related Works}
\label{sec:related_works}
We classify defence methods into two categories based on the underlying technical approach: clustering-alike defence and robust training-based defence. 

\subsection{Clustering Alike Defence}
This category reviews works that determine whether a node is malicious based on its relevance to the majority, using techniques such as cosine similarity and machine learning model interpretability to defend against malicious nodes, similar to clustering. \cite{tolpeginDataPoisoningAttacks2020} investigates how malicious participants can compromise federated learning by introducing poisoned data. Their study focuses on label-flipping attacks, where attackers modify class labels in their local datasets to manipulate the global model. They propose a defence mechanism using principal component analysis (PCA) to detect anomalous updates from malicious participants, demonstrating its effectiveness in isolating such participants. MOAT~\cite{manna2021moat} analyses the interpretability of machine learning models based on SHAP (i.e., Shapley Additive Explanations) features~\cite{NIPS2017_7062}. It posits that malicious and benign models have different interpretabilities, as SHAP features reveal these differences. However, MOAT relies on Z-scores and a threshold to determine whether a model is malicious, which may not be consistent across all datasets, as pointed out by SHERPA~\cite{sandeepaSHERPAExplainableRobust2024}. To address this issue, SHERPA improves MOAT by incorporating the HDBSCAN clustering method and a scoring mechanism that is less sensitive to the variability of different datasets. CONTRA~\cite{awanCONTRADefendingPoisoning2021} employs a cosine-similarity-based measure to assess the credibility of local model parameters, alongside a reputation scheme to dynamically adjust the contributions of each node. This method assumes that benign nodes will be more similar to the majority compared to malicious nodes. Extensive experiments show that CONTRA significantly reduces the attack success rate and minimises degradation of global model performance compared to state-of-the-art defences. MCDFL~\cite{jiangDataQualityDetection2023}, which stands for Malicious Clients Detection Federated Learning, proposes a defence mechanism against label-flipping attacks in federated learning. By recovering a distribution over a latent feature space, MCDFL can identify malicious nodes and assess the data quality of each node. It assumes that benign nodes will exhibit similar data quality to the majority, while malicious nodes will not. The proposed strategy is tested on benchmark datasets such as CIFAR10 and Fashion MNIST, using different neural network models and attack scenarios. Results show that this solution is robust in detecting malicious nodes without incurring excessive costs. VFedAD~\cite{laiVFedADDefenseMethod2023}, meaning a Defence Method Based on the Information Mechanism Behind Vertical Federated Data Poisoning Attacks, addresses the challenges posed by vertical federated data poisoning attacks. It proposes an unsupervised defence method rooted in information theory that learns semantic-rich representations of client data to effectively detect anomalies, thereby protecting subsequent algorithms from attacks. Similar to MCDFL, it assumes that benign nodes will have data representations similar to the majority, while malicious nodes will not. Experimental results validate VFedAD’s ability to detect anomalies introduced by data poisoning attacks, demonstrating its effectiveness as a defence mechanism. However, these methods may struggle in scenarios with a high ratio of malicious nodes, as a significant percentage of malicious nodes can dominate and distort the standard used for comparison among nodes.

\subsection{Robust Training Based Defence}
This category focuses on studies that utilise machine learning model aggregation strategies and noise injection techniques to mitigate the impact of malicious nodes. Krum~\cite{blanchard_machine_2017} proposed an aggregation rule (i.e., a method for merging local machine learning models) that first calculates the squared sum of distances between the weight vector of the current node’s model and the weight vectors of the $n-f-2$ closest nodes. The rule then selects the weight vector with the smallest squared sum distance as the global model's weight vector. The work by Yin et al.~\cite{yin_byzantine-robust_2018} introduced a median-based merging method. This approach uses the coordinate-wise median value to replace the average value used in~\cite{mcmahan2017communication} when merging local models into a global model. Trimmed Mean~\cite{yinByzantineRobustDistributedLearning2018} proposes a method for trimming the model weight values based on a predefined trim parameter $K$, with the condition that $k < \frac{n}{2}$, where $n$ is the number of nodes. The selected model weights, determined by the trim parameter, are then aggregated into a global model using a mean aggregation scheme. FRIENDS~\cite{liuFriendlyNoiseAdversarial2022} is a noise injection defence mechanism, short for Friendly Noise Defence, designed to counter data poisoning attacks in deep learning. The authors identify that poisoning attacks create local sharp regions with high training loss, which facilitate adversarial perturbations. To neutralise this, FRIENDS employs a two-part noise strategy: (1) Friendly noise, which introduces optimised perturbations that do not degrade model performance but disrupt attack success, and (2) Random noise, which varies across training iterations to prevent adaptive attacks from overcoming the defence. FRIENDS effectively defends against major poisoning attacks while maintaining high model accuracy and minimal computational overhead. It is also transferable across different architectures, making it practical for real-world deep learning applications. However, the introduction of noise adds additional complexities during training and inevitably affects the model's accuracy.

Existing methods for detecting malicious nodes have a significant limitation: their effectiveness declines when the proportion of malicious nodes exceeds a certain threshold. For instance, clustering-alike defence methods become ineffective when more than 50\% of the group consists of malicious nodes, as these methods primarily focus on the majority class. When the majority of nodes are malicious, the algorithms struggle to accurately classify which nodes are benign and which are malicious. Similarly, robust training methods can be overwhelmed when the majority of nodes are malicious.

\section{Conclusion}
\label{sec:conclusion}
This paper described the FedLAD data poisoning defence method for FL. FedLAD exhibits high tolerance to malicious nodes in terms of data poisoning in an FL system, thereby providing greater robustness. Comprehensive experiments demonstrated the advantages of our method with various malicious settings. While this paper applies FedAVG on the RREF pivots, our future work will explore other aggregation schemes such as Median, Krum, etc. to study further possible improvements of the robustness of FedLAD.

%%%%%%%%%%%%%%%%%%%%%%%%%%%%%%%%%%%%%%%%%%%%%%%%%%%%%%%%%%%%%%%%%%%%%%%%

%%% Use this command to include your bibliography file.

\bibliography{main.bib}

\end{document}